\pdfoutput=1
\documentclass[11pt]{article}

\usepackage[]{acl}

\usepackage{times}
\usepackage{latexsym}
\usepackage{booktabs}
\usepackage{multirow}
\usepackage{subcaption}
\usepackage{tcolorbox}
\usepackage{multirow, colortbl, caption}
\usepackage[T1]{fontenc}

\usepackage[utf8]{inputenc}

\usepackage{microtype}

\usepackage{inconsolata}

\usepackage{graphicx}

\title{Judging the Judges: Can Large Vision-Language Models Fairly Evaluate Chart Comprehension and Reasoning? }
 \author{Md Tahmid Rahman Laskar\textsuperscript{\textdaggerdbl,}\thanks{\hspace{0.115cm} Contact Emails: \{tahmedge,enamulh,jhuang\}@yorku.ca},  \textbf{Mohammed Saidul Islam\textsuperscript{\textdaggerdbl ,}\thanks{\hspace{0.115cm} Equal Contributions.}}\textbf{,}   \textbf{Ridwan Mahbub\textsuperscript{\textdaggerdbl ,}\footnotemark[2]}\textbf{,} \\  \textbf{Ahmed Masry\textsuperscript{\textdaggerdbl}}\textbf{,}  \textbf{Mizanur Rahman}\textsuperscript{\textdaggerdbl}\textbf{,}   %\textbf{,} 
\textbf{Md Amran Hossen Bhuiyan\textsuperscript{\textdaggerdbl}}\textbf{,} \\
 \textbf{Mir Tafseer Nayeem\textsuperscript{\textsection}}\textbf{,}  %\textbf{,} 
 \textbf{Shafiq Joty\textsuperscript{\textparagraph,\textdollar}}\textbf{, }
 \textbf{Enamul Hoque\textsuperscript{\textdaggerdbl,}}\footnotemark[1]\textbf{, }
 \textbf{Jimmy Xiangji Huang\textsuperscript{\textdaggerdbl,}\footnotemark[1]} \\
            {\textsuperscript{\textdaggerdbl}York University, 
                     \textsuperscript{\textsection}University of Alberta, 
\textsuperscript{\textparagraph}Nanyang Technological University,
\textsuperscript{\textdollar}Salesforce AI Research}
          \\ 
          } 

\begin{document}
\maketitle
\begin{abstract}
Charts are ubiquitous as they help people understand and reason with data. Recently, various downstream tasks, such as chart question answering, chart captioning, etc. have emerged. Large Vision-Language Models (LVLMs) show promise in tackling these tasks, but their qualitative evaluation is costly and time-consuming, limiting real-world deployment. While using LVLMs as judges to assess chart comprehension capabilities of other LVLMs could streamline evaluation processes,  challenges like proprietary datasets, restricted access to powerful models, and evaluation costs hinder their adoption in industrial settings. To this end, we present a comprehensive evaluation of 13 open-source LVLMs as judges for diverse chart comprehension and reasoning tasks. We design both pairwise and pointwise evaluation tasks covering criteria like factual correctness, informativeness, and relevancy. Additionally, we analyze LVLM judges based on format adherence, positional consistency, length bias, and instruction-following.  We focus on cost-effective LVLMs ($\leq9$B parameters) suitable for both research and commercial use, following a standardized evaluation protocol and rubric to measure the LVLM judge accuracy. Experimental results reveal notable variability: while some open LVLM judges achieve GPT-4-level evaluation performance (about 80\% agreement with GPT-4 judgments), others struggle (below ~10\% agreement). Our findings highlight that state-of-the-art open-source LVLMs can serve as cost-effective automatic evaluators for chart-related tasks, though biases such as positional preference and length bias persist.
\end{abstract}
\begin{figure}[t]
    \centering
    \includegraphics[width=\columnwidth]
    {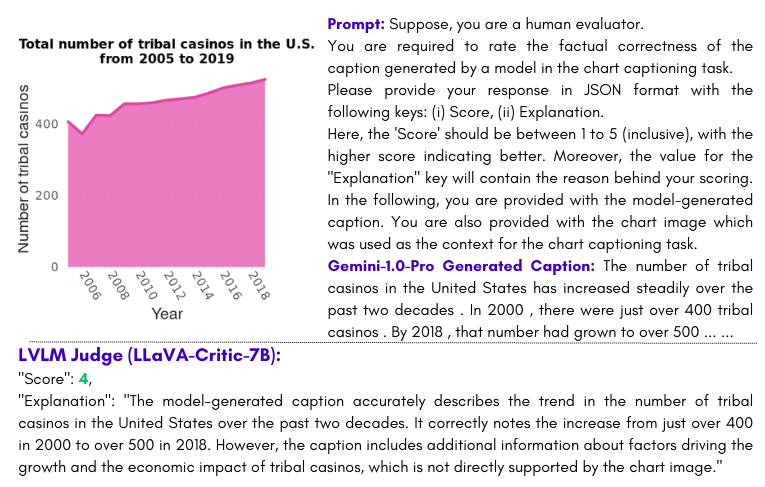}
    \caption{An example evaluation of \textit{Gemini-1.0-Pro} model generated L2/L3 caption in the VisText dataset by an LVLM judge: the \textit{LLaVA-Critic-7B} model. } %Here, `\dots' is used for brevity.}
    %\vspace{-1mm}
    \label{fig:intro_figure}
\end{figure}

%\vspace{-1mm}
\section{Introduction}
\newcommand{\Sref}[1]{\S\ref{#1}}
\newcommand{\sref}[1]{\S\ref{#1}}

%\vspace{-1mm}
%Chart data requires nuanced multimodal reasoning to interpret \cite{huang2024pixels}. 
% Data visualizations such as bar and line charts are widely used for data interpretation and multimodal reasoning 
Understanding data visualizations—such as bar and line charts—requires multimodal reasoning, as it involves integrating visual encodings with textual and contextual information~\cite{hoque2024natural}. 
Recent research has introduced various tasks (e.g., chart question answering, chart captioning, fact-checking with charts, etc.) to facilitate chart-based reasoning via natural language. These tasks demand the understanding of both the chart’s visual content (data values, trends, visual encodings) and accompanying text or instructions. % These include (not limited to) tasks like chart question answering, chart summarization, and fact-checking with charts.  %Large Language Models (LLMs) have revolutionized NLP and vision-language tasks, yet their abilities in data visualization domains remain under-explored. This gap is partly due to the specialized nature of charts – effective reasoning requires multi-modal capabilities beyond text, combining vision and data comprehension.

Large Language Models (LLMs) have revolutionized NLP and vision-language tasks \cite{zhao2023survey}, with growing interest in their use for chart comprehension and reasoning due to their strong multimodal capabilities. %. Chart comprehension and reasoning through LLMs have gained significant popularity due to their impressive capabilities in interpreting visual data. 
This progress can have a substantial impact on real-world industrial applications, where extracting insights from charts and graphs can drive critical business decisions~\cite{obeid-hoque-2020-chart,masry-etal-2023-unichart,meng2024chartassistant}. However, evaluating LLM performance in chart understanding presents notable challenges \cite{islam2024large}. For instance, traditional text-based metrics like BLEU fail to capture the quality of open-ended explanatory answers and also require human-annotated references. While human evaluation can address this problem, it is time-consuming and resource-intensive.

%An appealing alternative is LLM-as-a-Judge – using an LLM to automatically evaluate model outputs based on certain criteria
%If reliable, such automated judges can greatly accelerate development in this space by providing consistent, reproducible evaluations without human intervention.
To address this, recent studies have proposed using LLMs themselves as automatic evaluators or judges \cite{gu2024survey,li2024llms}. By employing LLMs to evaluate the chart comprehension abilities of other models (see Figure \ref{fig:intro_figure} for an example), the evaluation process can be streamlined, making the process more efficient and reproducible without human intervention. %While this approach could enable faster iteration cycles for improving AI systems and allow industries to more readily incorporate advanced chart reasoning capabilities into their workflows, there are some restrictions in real-world industrial scenarios that prohibit the use of many existing powerful LLMs. 
% While this approach can accelerate model development and support industry adoption of advanced chart reasoning, real-world deployment faces challenges.
% For instance, many industries may not share their proprietary charts and data with the closed-source model providers (e.g., OpenAI, Google, Antrhopic, etc.)
While this method accelerates development and reduces dependency on human annotations, its real-world adoption is hindered by privacy and scalability constraints. For example, organizations may be unwilling to share proprietary data with closed-source models from OpenAI, Google, or Anthropic. While closed-source models demonstrate impressive judging capabilities, their compatible open-source models are often large in size (e.g., 70B to 400B parameters). This requires high computing resources and usage costs. Therefore hinders real-world utilization. 

To this end, this paper aims to investigate whether open-source smaller LVLMs (e.g., less than 10B parameters) can effectively evaluate answers about charts---assessing correctness, relevance, and other qualities---similarly to a human or a powerful LLM like GPT-4 \cite{openai2023gpt4}. For this purpose, we conduct one of the first comprehensive evaluations of open-source LVLMs as evaluators on various chart benchmarks, consisting of diverse tasks like chart captioning and question answering. We focus on open-source, smaller VLMs (up to 10B parameters) to simulate realistic deployment scenarios where cost-effective or private models are preferred over large closed models. By benchmarking these models against high-quality reference judgments generated by closed-source LLMs like GPT-4 or 70B open-source LLM-Judge like LLaVA-Critic \cite{xiong2024llavacritic}, we aim to uncover to what extent current open models can serve as reliable judges, and when they fail.

Our major contributions to this paper are:
\begin{enumerate}
    \item We establish an evaluation framework for chart comprehension using \emph{``LVLM-as-a-Judge''}, with clear rubrics for pairwise and pointwise assessments over 100K judgments generated by GPT-4o and LLaVA-Critic-70B. Additionally, we introduce a new benchmark to assess the instruction-following abilities of LVLMs in chart-related tasks.
    \item We evaluate a wide range of open-source multimodal LLMs as judges – 13 models ranging from 2B to 9B parameters – and analyze their performance against LLM-annotated (GPT-4 and LLaVA-Critic) and human-annotated reference judgments, across diverse chart benchmarks (OpenCQA and VisText) on answers generated by different LLMs to create challenging evaluation scenarios. 
    \item We provide an in-depth analysis of the judges’ strengths and weaknesses, revealing issues like position bias and length bias, and discuss which models achieve substantially higher agreement with reference judgments, and which ones fail. 
\end{enumerate}

In addition, our code, judgment data, and our proposed instruction-following evaluation benchmark is released here: \url{https://github.com/tahmedge/chart_lvlm_judge}

%\vspace{-1mm}

\section{Related Work}
%\vspace{-1mm}
%Earlier efforts in chart question answering include synthetic datasets like FigureQA \cite{kahou2017figureqa} and DVQA \cite{dvqa}, which generated templated QA pairs for simple charts. However, these datasets lacked the complexity of real-world charts. The ChartQA \cite{masry-etal-2022-chartqa} benchmark is a noticeable exception in this regard, consisting of real-world charts with expanded question complexity and data sources. The OpenCQA \cite{open-CQA} benchmark further advanced the field by introducing open-ended questions that require explanatory answers about real-world charts. In parallel, chart captioning has been explored as a way to summarize chart content \cite{obeid-hoque-2020-chart, chart-to-text-acl,Rahman_2023,2023-vistext}. All these datasets highlight the key challenge that models must combine visual decoding with reasoning for generating responses related to the charts. 

Earlier efforts in chart question answering include synthetic datasets like FigureQA \cite{kahou2017figureqa} and DVQA \cite{dvqa}, which generated templated QA pairs for simple charts but lacked real-world complexity. ChartQA~\cite{masry-etal-2022-chartqa} addressed this gap with real-world charts and more complex questions, while OpenCQA \cite{open-CQA} pushed further with open-ended, explanatory queries. Meanwhile, chart captioning has emerged as another avenue for summarizing chart content \cite{chart-to-text-acl,Rahman_2023,2023-vistext}. Together, these datasets highlight the growing complexity of chart-based reasoning tasks and the need for more robust evaluation methods.
% Collectively, these datasets underscore the challenge of combining visual decoding with reasoning to generate meaningful chart-related responses.

%While the rise of multimodal LLMs opens the possibility of tackling chart tasks using them, the general vision-language models often struggle with chart-specific content like reading axis text or precise data points \cite{islam2024large}. Therefore, specialized models like ChartLLaMA \cite{han2023chartllama}, ChartInstruct \cite{masry2024chartinstruct}, ChartGemma \cite{masry2025chartgemma}, TinyChart ~\citep{zhang2024tinychartefficientchartunderstanding} etc. are proposed that achieved impressive performance in performing chart-related tasks. Despite these advances, evaluating such models remains challenging. Many of these works still rely on human evaluation to judge open-ended responses, which is time-consuming and not scalable.

While the rise of multimodal LLMs offers potential for chart-related tasks,  general vision-language models often struggle with chart-specific elements like axis text and precise data points~\cite{islam2024large}. To address this, specialized models such as ChartLLaMA \cite{han2023chartllama}, ChartInstruct \cite{masry2024chartinstruct}, ChartGemma~\cite{masry2025chartgemma}, and TinyChart \citep{zhang2024tinychartefficientchartunderstanding} have been developed, showing strong performance. However, evaluating these models is challenging, as many still depend on time-consuming %, non-scalable 
human assessments for open-ended responses.

While using LLMs to evaluate other LLMs has gained a lot of attention, early efforts focused primarily on text-only tasks like summarization \cite{li2024llms, zheng2023judging}. For multimodal tasks, models such as Prometheus-VL \cite{lee2024prometheusvl} and LLaVA-Critic \cite{xiong2024llavacritic} introduced smaller open-source vision-language models (as small as 7B) fine-tuned to serve as general-purpose multimodal evaluators. Our work aligns with this direction, leveraging LVLMs as judges. Although concurrent studies explore similar capabilities \cite{chen2024mllm}, they report that early LVLMs like LLaVA-1.5 struggle with text-rich visuals such as charts and diagrams \cite{lee2024prometheusvl}. Addressing the gap in evaluating recent LVLMs on chart-specific tasks, we present the first systematic study of state-of-the-art open-source LVLMs as judges across diverse chart comprehension and reasoning benchmarks.

\begin{figure}[t]
    \centering
    \includegraphics[width=\columnwidth]{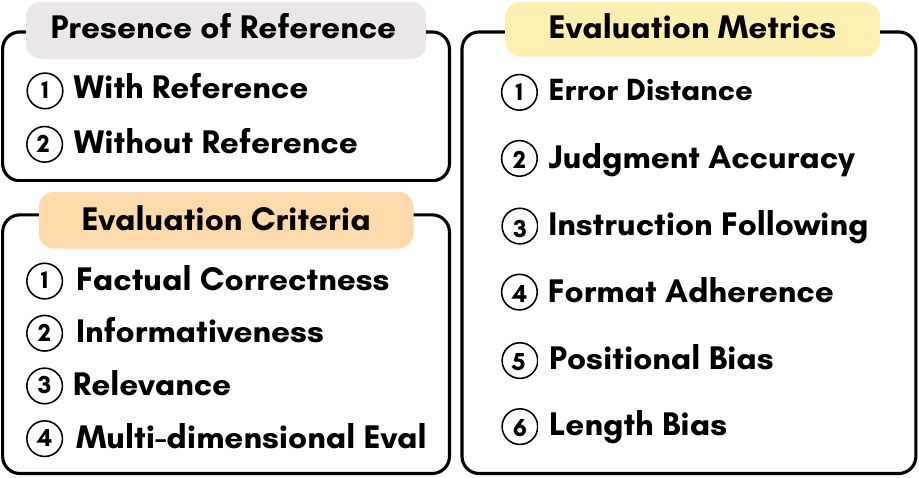}
    \caption{An overview of our evaluation methodology.}
    %\vspace{-1mm}
    \label{fig:charteval_method}
\end{figure}

%\vspace{-1mm}
%  (see Section \ref{rubric_design})
\section{Methodology}
%\vspace{-1mm}
Given a chart and model generated response(s), we construct the prompt (see Appendix \ref{appendix_prompt} for some sample prompts) depending on the evaluation rubric. Following the prior work \cite{lee2024prometheusvl}, we ask the LVLM judge to provide their answer along with an explanation, since adding an explanation during assessments ensured better judgment performance in early work. Below, we describe our evaluation method (also see Figure \ref{fig:charteval_method}). 
%\vspace{-1mm}
\subsection{Evaluation Rubric Design} 
\label{rubric_design}
Following the prior work on LLM evaluation \cite{chen2024mllm,lee2024prometheusvl,xiong2024llavacritic}, we define clear rubrics for the judges:

\noindent \textit{(i) Based on Evaluation Type:}
 
    - \textbf{Pairwise:} The judge must select the better answer between two given responses (e.g., Claude vs Gemini)  about the chart for the given instruction.
     
     - \textbf{Pointwise:} The judge must rate a single answer to the chart query on a Likert scale from 1 (very poor) to 5 (excellent).

\noindent \textit{(ii) Based on Reference Type:}

    - \textbf{With Reference:} The judge is also given the ground-truth answer or summary as a reference, and instructed to choose the response that better matches the reference as well as the chart context.
       
        - \textbf{Without Reference:} The judge only sees the model response(s) and the chart image and must decide based on its own judgment.
   
\noindent \textit{(iii) Based on Evaluation Criteria:}
  
   - \textbf{Factual Correctness:} Focuses only on the factual accuracy of the response.

        - \textbf{Informativeness:} Focuses on the amount of useful information in the response. 

- \textbf{Relevance:} Focuses on measuring the relevancy of the response. 
 
 - \textbf{Multidimensional Evaluation:} Considers overall response quality based on factual correctness, informativeness, conciseness, and relevance.

\noindent \textit{(iv) Based on Evaluation Metrics:}
  
   - \textbf{Judgment Accuracy:}  The percentage of instances where the answer picked by the judge same as the gold. It is relevant to the pairwise case. 
     
               - \textbf{Error Distance:} The average absolute difference between the judge’s 1–5 rating and the reference’s rating. It is relevant to the pointwise case. 
   
        - \textbf{Positional Bias Metric:} In the pairwise case, we swapped the order of answers and checked if the judge’s decision changed. 
    
       - \textbf{Length Bias Metric:} Checked if the judge’s wrong choice correlated with the answer length.

       - \textbf{Instruction Following Evaluation Accuracy:} We analyzed whether the LVLM judge can effectively evaluate the instruction following capability of other LVLMs.
        
        - \textbf{Format Adherence Accuracy:} This metric measures whether the judge’s output followed the required JSON format. 
        
%\vspace{-1mm}
\subsection{Evaluation Data Construction}
%\vspace{-1mm}
\textbf{OpenCQA} \cite{open-CQA}\textbf{:} This is an open-ended question-answering dataset on real charts. Each data point includes a chart and a question and expects an explanatory answer. We use its test set containing 1.1k QA instances.

\noindent \textbf{VisText} \cite{2023-vistext}\textbf{:} This is a chart captioning dataset with 12,441 charts, each paired with two types of captions: synthetic Level 1 (L1) captions that describe the chart's structural elements—such as chart type, title, axis labels, and scales—and human-generated Level 2/Level 3 (L2/L3) captions that provide insights into key statistics, trends, and patterns within the data.  We use both L1 and L2/L3 captions with 1.2K test instances for each type. 

For OpenCQA and VisText, we use the outputs generated by \citet{islam2024large} using \textit{Gemini-1.0-Pro}  \cite{geminiteam2023gemini} and \textit{Claude-3-Haiku} \cite{Claude} and compute the judgment scores using GPT-4o \cite{openai2023gpt4} and LLaVA-Critic-70B \cite{xiong2024llavacritic} models and use as the judgment reference for diverse scenarios, as demonstrated in the previous section. This results in about 100K judgment data generated by GPT-4o and LLaVA-Critic-70B. We select these two models due to their impressive performance as a multimodal LLM-Judge \cite{xiong2024llavacritic}. 

\noindent \textbf{Chart-Instruct-Eval:} We find that there are no datasets currently available in the chart domain that can assess the instruction-following capabilities of LVLMs. Therefore, we construct an instruction-following dataset (denoted as Chart-Instruct-Eval) to evaluate whether LVLM judges can evaluate the instruction-following capabilities of different models in chart-related tasks. For the dataset construction, we sample 400 charts from the ChartGemma \cite{masry2025chartgemma} dataset. % and create {detailed instructions for each sample replacing the original input instructions 
However, the original input instructions in the ChartGemma dataset lacked sufficient details. Hence, we could not use it for the instruction following purpose. Therefore, for each sample, we first create a detailed instruction containing specific requirements for the LLM response in terms of formatting, length, and structure to ensure instruction following. Then we manually prepare one good and one bad response corresponding to the instruction. Both responses convey similar content, but the good response fully adheres to all provided instructions, whereas the bad response disregards them. Finally, we assess the LLM judges whether they can reliably evaluate which response properly follows the instructions.

%\vspace{-1.5mm}
\subsection{LVLM Judges}
%\vspace{-1mm}
We evaluate \textbf{13} different open-source multimodal LLMs\footnote{We did not use the Prometheus-VL-7B \cite{lee2024prometheusvl} model since it requires a specific input format, making our prompts incompatible. } as candidate judges, focusing on relatively smaller, publicly available models (2B–10B parameters). These include:
% \footnote{\url{https://huggingface.co/meta-llama/Llama-3.2-11B-Vision}
%\textbf{(i)} \texttt{LLaMA-3.2-11B-Vision-Instruct} – \textit{a vision-enhanced LLaMA-3 \cite{grattafiori2024llama3} model developed by Meta},
\textbf{(i)} \texttt{XGen-MM-Phi-3-3.8B}~\cite{xue2024xgen} – \textit{a multimodal model (3.8B) developed by Salesforce}, 
\textbf{(ii)} \texttt{MiniCPM-V-2.6-7B} \cite{yao2024minicpm} – \textit{a 7B vision-language model by OpenBMB}, \textbf{(iii)} \texttt{Ph-3.5-3.8B-Vision-Instruct}  \cite{abdin2024phi} – \textit{a smaller vision model from Microsoft}, \textbf{(iv)} \texttt{Qwen2-VL-2B}  - \textit{Alibaba’s Qwen \cite{wang2024qwen2} multimodal model with just 2B parameters}, \textbf{(v)} \texttt{Qwen2-VL-7B} – \textit{The 7B version of the multimodal Qwen model}, \textbf{(vi)} \texttt{PaliGemma-3B}~\cite{beyer2024paligemma} – \textit{Google’s multimodal open-source model}, \textbf{(vii)} \texttt{ChartGemma} \cite{masry2025chartgemma} – \textit{a chart-specialized model based on PaliGemma that is fine-tuned on chart tasks}, \textbf{(viii)} \texttt{Idefics-9B-Instruct\footnote{\href{https://huggingface.co/HuggingFaceM4/idefics-9b-instruct}{HuggingFaceM4/idefics-9b-instruct}}} – \textit{an open multimodal model known for image understanding}, \textbf{(ix)} \texttt{InternLM-XComposer-7B} \cite{dong2024internlm} – \textit{a 7B vision model with composition abilities}, \textbf{(x)} \texttt{LLaVA-v1.6-Mistral-7B} – \textit{A multimodal LVLM based on the LLaVA \cite{li2024llava} architecture that also utilizes a 7B Mistral \cite{jiang2023mistral7b} as the backbone}, \textbf{(xi)} \texttt{LLaVA-Critic-7B} – \textit{a specialized evaluator model based on LLaVA and Qwen}, \textbf{(xii)} \texttt{mPLUG-Owl-3-7B} \cite{ye2023mplug} – \textit{a 7B multimodal model from  Alibaba}, \textbf{(xiii)} \texttt{Janus-Pro-7B}~\cite{chen2025janus} - \textit{an open-source LVLM developed by Deepseek}. For more information about model selection, see Appendix \ref{appendix:data_model_selection_criteria}. %, and \textbf{(xv)} \texttt{Prometheus-Vision }\cite{lee2024prometheusvl} -  \textit{an open-source fine-tuned multimodal LLM judge}. 

% \footnote{\url{https://huggingface.co/liuhaotian/llava-v1.6-mistral-7b}}

%\vspace{-1mm}

\section{Experiments}
%\vspace{-1mm}
\definecolor{open_models_below_4B}{RGB}{185, 235, 255}
\definecolor{open_models_7B_12B}{RGB}{255, 219, 187}

\begin{table*}[t!]
    \scriptsize
    \setlength{\tabcolsep}{3pt}
    \resizebox{\textwidth}{!}{
        \centering
        \begin{tabular}{c|ccc|ccc|ccc|ccc|ccc|ccc}
            \toprule
            \multicolumn{1}{c}{} &
              \multicolumn{9}{c}{\textbf{Pairwise Evaluation: Judgment Accuracy (Higher is Better)}} &
              \multicolumn{9}{c}{\textbf{Pointwise Evaluation: Error Distance (Lower is Better)}}\\
            \cmidrule(lr){2-10}\cmidrule(lr){11-19}
            \multicolumn{1}{c}{\multirow{2}{*}{\textbf{Model}}} &
              \multicolumn{3}{c}{\textbf{OpenCQA}} &
              \multicolumn{3}{c}{\textbf{VisText L1}} &
              \multicolumn{3}{c}{\textbf{VisText L2/L3}} &
              \multicolumn{3}{c}{\textbf{OpenCQA}} &
              \multicolumn{3}{c}{\textbf{VisText L1}} &
              \multicolumn{3}{c}{\textbf{VisText L2/L3}}\\
            \cmidrule(lr){2-4}\cmidrule(lr){5-7}\cmidrule(lr){8-10}\cmidrule(lr){11-13}\cmidrule(lr){14-16}\cmidrule(lr){17-19}
            \multicolumn{1}{c}{} &
              \textbf{GPT‑4o} & \textbf{LC‑70B} & \textbf{Avg.} &
              \textbf{GPT‑4o} & \textbf{LC‑70B} & \textbf{Avg.} &
              \textbf{GPT‑4o} & \textbf{LC‑70B} & \textbf{Avg.} &
              \textbf{GPT‑4o} & \textbf{LC‑70B} & \textbf{Avg.} &
              \textbf{GPT‑4o} & \textbf{LC‑70B} & \textbf{Avg.} &
              \textbf{GPT‑4o} & \textbf{LC‑70B} & \textbf{Avg.}\\
            \midrule
\rowcolor{open_models_below_4B!50} Qwen2‑VL‑2B‑Instruct
            & 51.6 & 56.3 & 54.0
            & 28.5 & 25.9 & 27.2
            &  2.5 &  3.4 &  3.0
            &  1.0 &  0.9 &  1.0
            &  2.0 &  2.1 &  2.1
            &  1.1 &  0.6 &  0.9\\
\rowcolor{open_models_below_4B!50} PaliGemma‑3B
            & 0.0 & 0.0 & 0.0
            & 0.0 & 0.0 & 0.0
            & 0.0 & 0.0 & 0.0
            & 5.0 & 5.0 & 5.0
            & 5.0 & 5.0 & 5.0
            & 5.0 & 5.0 & 5.0\\
\rowcolor{open_models_below_4B!50} ChartGemma‑3B
            & 0.0 & 0.0 & 0.0
            & 0.0 & 0.0 & 0.0
            & 0.0 & 0.0 & 0.0
            & 5.0 & 5.0 & 5.0
            & 5.0 & 5.0 & 5.0
            & 5.0 & 5.0 & 5.0\\
\rowcolor{open_models_below_4B!50} Phi‑3.5‑Vision‑3.8B‑Instruct
            & 49.5 & 51.9 & 50.7
            & 72.5 & 66.4 & 69.5
            & 43.6 & 55.3 & 49.5
            &  0.7 &  0.8 &  0.8
            &  1.4 &  1.6 &  1.5
            &  1.1 &  0.9 &  1.0\\
\rowcolor{open_models_below_4B!50} XGen‑MM‑Phi3‑3.8B‑Instruct
            & 67.6 & 75.5 & 71.6
            & 78.5 & 72.2 & 75.4
            & 63.9 & 77.4 & 70.7
            &  1.0 &  0.7 &  0.9
            &  1.3 &  1.5 &  1.4
            &  1.0 &  0.4 &  0.7\\
\rowcolor{open_models_7B_12B!50} Janus‑Pro‑7B
            & 46.6 & 48.7 & 47.7
            & 48.6 & 45.6 & 47.1
            & 52.6 & 57.0 & 54.8
            &  1.0 &  0.7 &  0.9
            &  1.0 &  1.2 &  1.1
            &  1.0 &  0.4 &  0.7\\
\rowcolor{open_models_7B_12B!50} Qwen2‑VL‑7B‑Instruct
            & 67.3 & 66.4 & 66.9
            & 64.0 & 51.1 & 57.6
            & 69.6 & 70.3 & 70.0
            &  0.8 &  0.6 &  0.7
            &  0.6 &  {0.5} &  {0.6}
            &  0.9 &  0.5 &  0.7\\
\rowcolor{open_models_7B_12B!50} InternLM‑Xcomposer2d5‑7B
            & 64.8 & 64.1 & 64.5
            & 76.8 & 67.2 & 72.0
            & \textbf{69.7} & 81.4 & 75.6
            &  0.8 &  0.9 &  0.9
            &  0.8 &  0.9 &  0.9
            &  0.9 &  0.4 &  0.7\\
\rowcolor{open_models_7B_12B!50} LLaVA‑Next‑v1.6‑Mistral‑7B
            & 72.0 & 79.8 & 75.9
            & 78.4 & 71.7 & 75.1
            & 66.7 & 83.4 & 75.1
            &  0.9 &  0.6 &  0.8
            &  1.3 &  1.5 &  1.4
            &  1.1 &  0.6 &  0.9\\
\rowcolor{open_models_7B_12B!50} \textbf{LLaVA‑Critic‑7B}
            & \textbf{75.1} & \textbf{83.8} & \textbf{79.5}
            & \textbf{82.8} & \textbf{75.3} & \textbf{79.1}
            & 69.0 & \textbf{85.1} & \textbf{77.1}
            & \textbf{0.5} & \textbf{0.4} & \textbf{0.5}
            & \textbf{0.5} & \textbf{0.4} & \textbf{0.5}
            & \textbf{0.8} & \textbf{0.4} & \textbf{0.6}\\
\rowcolor{open_models_7B_12B!50} mPLUG‑Owl3‑7B
            & 60.8 & 59.4 & 60.1
            & 72.2 & 64.0 & 68.1
            & 46.1 & 39.2 & 42.7
            &  0.8 &  0.6 &  0.7
            &  1.0 &  1.0 &  1.0
            &  0.9 &  0.4 &  0.7\\
\rowcolor{open_models_7B_12B!50} MiniCPM‑V‑2.6‑8B
            & 64.3 & 68.6 & 66.5
            & 49.2 & 42.9 & 46.1
            & 44.8 & 39.1 & 42.0
            &  1.0 &  0.8 &  0.9
            &  1.3 &  1.3 &  1.3
            &  1.7 &  1.5 &  1.6\\
\rowcolor{open_models_7B_12B!50} Idefics‑9B‑Instruct
            & 20.4 & 20.1 & 20.3
            & 22.0 & 19.7 & 20.9
            & 24.1 & 24.4 & 24.3
            &  3.3 &  3.2 &  3.3
            &  4.8 &  4.8 &  4.8
            &  3.1 &  2.8 &  3.0 \\
            \bottomrule
        \end{tabular}
    }
    \caption{Model performance based on average pointwise and pairwise scores across all reference types, as well as evaluation criteria (e.g., factual correctness, informativeness, etc.) in comparison to GPT‑4o and LLaVA‑Critic‑70B (LC‑70B) annotations (corresponding average score is also added). Bold values denote the best score in each case. Color coding for comparison:
    open-source models \colorbox{open_models_below_4B!50}{below 7B parameters}, \colorbox{open_models_7B_12B!50}{between 7-10B parameters}.}
    \label{tab:overall_avg_model_results}
\end{table*}

In this section, we present the experimental results based on evaluating 13 LVLMs as judges across OpenCQA, VisText, and our proposed Chart-Instruct-Eval. The evaluation considers both pairwise and pointwise assessments, focusing on factual correctness, informativeness, relevance, positional bias, length bias, and instruction-following accuracy. We parse the LVLM-judge predicted judgments from their corresponding JSON-formatted responses using a parsing script \cite{laskar2023systematic,laskar-etal-2024-systematic,laskar-etal-2024-query}. If the parsing script cannot properly parse the judgment from the response, we consider the LLM-generated answer as wrong for the pairwise case and error distance of 5 for the pointwise case.  Note that we ran all our experiments using 1 A100 GPU with all the decoding parameters being set to the default values in HuggingFace \cite{wolf2019huggingface}. Below, we demonstrate our findings:

%\vspace{-1mm}

\subsection{Pairwise Evaluation Results}
\label{pairwise_results}
%\vspace{-1mm}
The pairwise evaluation measures how often the LVLM judges agree with GPT-4 or LLaVA-Critic-70B to select the better response in comparative assessments. We summarize the result in Table \ref{tab:overall_avg_model_results}. % summarizes the accuracy scores for different models. % across OpenCQA and VisText (L1 and L2/L3).

i. \textbf{Top-performing models:} LLaVA-Critic-7B achieved the highest agreement with reference judgments (above 75\% average accuracy in each dataset). %While other
Another similar-sized (7B) LLM, % models, such as InternLM-Xcomposer-7B and 
the LLaVA-Next-v1.6-Mistral-7B model also performed competitively by exceeding 70\% accuracy across each dataset. Interestingly, the XGen-MM model with just 3.8B parameters also achieved more than 70\% accuracy, making it a very suitable judge in resource-constrained scenarios. 

ii. \textbf{Lower-performing models:} Surprisingly, PaliGemma-3B and ChartGemma-3B %, and Prometheus-Vision-7B 
achieved 0\% agreement, indicating a poor alignment with reference judgments. Moreover, while the Qwen-2B model achieves decent performance in OpenCQA (above 50\% accuracy), it achieves quite poor performance in VisText, especially in the L2/L3 scenario (below 10\% accuracy).  More surprisingly, the largest LVLM in our evaluation, the Idefics-9B-Instruct model achieves average accuracy below 25\% in all datasets, highlighting its ineffectiveness as a judge.
Our manual analysis revealed that these models failed due to not following instructions properly while also generating the response in the wrong format (improper JSON outputs). For PaliGemma, since it is not an instruction-tuned model, its poor performance could be related to the lack of understanding of instructions. The poor performance behind ChartGemma could be related to its training data lacking instructions related to judging tasks, therefore leading to poor generalization. We demonstrate some error examples of these LVLMs in Appendix \ref{error_analysis}. % for some error examples.
% Prometheus required a specific input  format, making our prompts incompatible. 
%Our manual analysis demonstrates that the  Prometheus model could only accept the input in a specified format and so our instruction prompts were not applicable. In terms of other poorly performing models, they did not follow the instructions properly and failed to generate the response in the required JSON format. %Moreover, some models 
%also tend to generate the answers or captions from the given chart, instead of performing as a judge. 
  
%\vspace{-1mm}
\subsection{Pointwise Evaluation Results}
\label{pointwise_results}
%\vspace{-1mm}
This primarily measures the error distance between the ratings of the LVLM judge and the reference (GPT-4/LLaVA-Critic-70B) on a 1–5 Likert scale. % Lower error distances indicate higher alignment.

i. \textbf{Top-performing models:} Similar to the pairwise scenario, we find from Table \ref{tab:overall_avg_model_results} that LLaVA-Critic-7B again achieved the best performance in the pointwise scenario, achieving an error distance around 0.5. % below 1.0. % confirming its ability to approximate reference judgments closely. 
 Other models like InternLM-Xcomposer-7B and Qwen2-VL-7B-Instruct that achieve quite good performance in pairwise scenarios, also demonstrate less error distance in pointwise scenarios (error distance below 1.0). 
 %although InternLM-Xcomposer-7B achieves an error distance below 1.0, the error distance for the XGen-MM model is above 1.3. 
Some other top-performing models in the pointwise scenario are LLaVA-Next-v1.5-Mistral-7B and MiniCPM-V-2.6-8B, which also achieve an error distance below 1.0 in 2 out of the 3 datasets. %and Janus-Pro-7B with both models having error distance below 1.0. %, demonstrating that they are more effective in pointwise scenarios.

ii. \textbf{Lower-performing models:} 
Similar to the pairwise scenario, PaliGemma-3B and ChartGemma-3B again produced irrelevant outputs resulting in the highest error distances (5.0). Moreover, despite being the largest model in our evaluation, the Idefics-9B-Instruct model performs quite poorly with a high error (on average, above 3). % an error distance over $3.0$. %Moreover, Llama-3.2-11B-Vision-Instruct again exhibited high error distances (\verb|~|4.0 across datasets), indicating significant deviation from reference ratings.  %, leading to their exclusion from further analysis.
%\vspace{-1mm}
\subsection{Instruction and Format Adherence}
%\vspace{-1mm}
We also assess the LVLM judges on their ability to maintain a standardized response format and whether they can evaluate the instruction following capabilities of other models. Based on the results in Table \ref{tab:format_instruction_accuracy_comparison}, we find that all 7B models achieve more than 90\% format following capability. Smaller LVLMs like Qwen-2B and Phi-3.8B also achieve around 80\% format adherence. 

In terms of instruction following capability evaluation, we find that many LVLMs that could properly follow the format following requirement in their generated judgments for pairwise (\Sref{pairwise_results}) and pointwise  (\Sref{pointwise_results}) evaluations, surprisingly generate the response in the wrong format in this evaluation. This makes our original parsing script penalize most of the LVLM-generated judgments as wrong. Therefore, we rewrite the parsing script to make it more flexible in terms of format adherence of the LVLM judge, since for this evaluation, our focus was to evaluate whether LVLM judges can properly assess instruction-following capabilities of different models in downstream chart-related tasks. Therefore, format adherence and other capabilities of the LVLM judges were not the focus of this evaluation. Our experiments reveal that mPLUG-Owl3-7B (93.5\%) and Qwen2-VL-7B-Instruct (87.0\%) achieve the top two results in terms of evaluating the instruction-following capability of different LVLM generated responses. % in terms of reliably evaluating the instruction-following capability of other LLMs. 
Surprisingly, the LLaVA-Critic-7B model achieves only 45.5\% accuracy in this task. This may indicate that the training data of the LLaVA-Critic-7B model may not contain such data, leading to a quite poor generalization in this dataset. 

Moreover, PaliGemma-3B and ChartGemma-3B fail to follow the format requirements at all, and also unable to evaluate instruction following capability. Finally, the Idefics-9B-Instruct model, even with 9B parameters, achieves poor instruction and format following accuracy.
%\vspace{-1.5mm}
\subsection{Bias Analysis}
%\vspace{-1mm}
To assess potential biases in LVLM judges, we analyzed position bias (whether the order of the responses affects judgments) and length bias (whether longer responses are favored). Based on the result presented in Table \ref{tab:bias_result}, we find that the Qwen2-VL-7B-Instruct model 
exhibited the lowest positional bias and length bias. 
On the contrary, the LLaVA-Next-v1.6-Mistral-7B model showed very high bias in both scenarios, suggesting susceptibility to judge responses based on variations in the position of the responses as well as the length. Surprisingly, the LLaVA-Critic-7B model, which is the best-performing model in terms of judgment accuracy and error distance, demonstrates the highest length bias across all models, indicating a tendency to favor longer answers. We provide an example of the position bias in Figure \ref{fig:position_bias}, and an example of the length bias in Figure \ref{fig:length_bias}.

\begin{table}[t]
\scriptsize
\setlength{\tabcolsep}{1.5pt} 
    \centering
    \begin{tabular}{l|c|c}
        \toprule
        \textbf{Model} & \textbf{ Instruction Following} & \textbf{Format Adherence} \\
        \midrule
                \rowcolor{open_models_below_4B!50} Qwen2-VL-2B-Instruct & 13.5 & 78.9 \\
                  \rowcolor{open_models_below_4B!50} PaliGemma-3B  & 0.0 & 0.0 \\
                   \rowcolor{open_models_below_4B!50} ChartGemma-3B  & 0.0 & 0.0 \\
     \rowcolor{open_models_below_4B!50} Phi-3.5-Vision-3.8B-Instruct & 49.0 & 83.3 \\
  \rowcolor{open_models_below_4B!50} XGen-MM-Phi3-3.8B-Instruct & 72.5 & 97.6 \\
            \midrule
            %\rowcolor{open_models_7B_12B!50} Prometheus-Vision-7B & 0.0 & 0.0 \\
            \rowcolor{open_models_7B_12B!50} Janus-Pro-7B & 73.0 & 96.7 \\

            \rowcolor{open_models_7B_12B!50} Qwen2-VL-7B-Instruct & 87.0 & 98.6 \\
      \rowcolor{open_models_7B_12B!50} InternLM-Xcomposer2d5-7B  & 54.0 & 95.9 \\
              \rowcolor{open_models_7B_12B!50} LLaVA- Next-v1.6-Mistral-7B  & 27.0 & 98.9 \\
            \rowcolor{open_models_7B_12B!50} \textbf{LLaVA-Critic-7B} & 45.5 & \textbf{99.7} \\

       \rowcolor{open_models_7B_12B!50} \textbf{mPLUG-Owl3-7B}  & \textbf{93.5} & 98.9 \\
    
        \rowcolor{open_models_7B_12B!50} MiniCPM-V-2.6-8B & 54.5 & 90.3 \\
        \rowcolor{open_models_7B_12B!50} Idefics-9B-Instruct & 20.5 & 35.0 \\
        %\rowcolor{open_models_7B_12B!50} Llama-3.2-11B-Vision-Instruct & 16.0 & 15.87 \\

        \bottomrule
    \end{tabular}
     %\vspace{-1mm}
    \caption{Accuracy in terms of Instruction Following Evaluation (evaluated on Chart-Instruct-Eval) and Format Adherence (based on average across all datasets).}
    \label{tab:format_instruction_accuracy_comparison}
\end{table}

\begin{table}[t]
\scriptsize
    \centering
    \begin{tabular}{l|c|c}
        \toprule
        \textbf{Model} & \textbf{Length Bias} & \textbf{Position Bias} \\
        \midrule
            \rowcolor{open_models_below_4B!50} Qwen2-VL-2B-Instruct & 55.1 & 71.9 \\
          
            \rowcolor{open_models_below_4B!50} Phi-3.5-Vision-3.8B-Instruct & 69.8 & 59.6\\
            
         \rowcolor{open_models_below_4B!50} XGen-MM-Phi3-3.8B-Instruct  & 64.3 & 79.2 \\
         \midrule
         \rowcolor{open_models_7B_12B!50} Janus-Pro-7B & 27.2 & 50.6 \\
          \rowcolor{open_models_7B_12B!50} \textbf{Qwen2-VL-7B-Instruct} & \textbf{21.5} & \textbf{35.8} \\
         \rowcolor{open_models_7B_12B!50} InternLM-Xcomposer2d5-7B & 24.5 & 35.9 \\
         \rowcolor{open_models_7B_12B!50} mPLUG-Owl3-7B & 21.9 & 42.5 \\
         \rowcolor{open_models_7B_12B!50} LLaVA-Next-v1.6-Mistral-7B & 71.8 & 77.0 \\
         \rowcolor{open_models_7B_12B!50} LLaVA-Critic-7B & 76.4 & 39.6 \\
    
        \rowcolor{open_models_7B_12B!50} MiniCPM-V-2.6-8B & 37.4 & 45.5 \\
        \bottomrule
    \end{tabular}
    %\vspace{-1mm}
    \caption{Length Bias and Position Bias for different models (results based on average across all datasets). Here, Lower values are better. Models achieving  format following accuracy above 50\% are only evaluated.}
    \label{tab:bias_result}
    %\vspace{-1mm}
\end{table}
%\vspace{-1mm}

\subsection{Human Evaluation}
%\vspace{-1mm}
In this section, we conduct a human evaluation of the GPT-4o and the LLaVA-Critic-70B models which we used as the reference judge to evaluate the smaller open-source LVLMs. For this purpose, we randomly sample 100 responses generated by \citet{islam2024large} for the Claude-3-Haiku and the Gemini-1-Pro models in OpenCQA and VisText datasets. Then, we ask two human annotators having expertise in NLP and Computer Vision to rate these responses based on our evaluation criteria (e.g., informativeness, relevance, etc.) with references provided for 50\% of the data and without any references for rest of the data. 

Based on our human evaluation, we find that both annotators' judgments highly correlate with GPT-4o and LLaVA-Critic-70B, with an error distance below 1.0. Interestingly, we find that both annotators have a higher correlation with the open-source LLaVA-Critic-70B model (average error distance with LLaVA-Critic-70B: \texttt{0.81}, and with GPT-4o: \texttt{0.93}). Therefore, in real-world industrial scenarios where human annotation is costly and closed-source LLMs are not preferred due to privacy concerns in proprietary datasets, the open-source LLaVA-Critic-70B model could be a good alternative for data annotation. % of the gold references to evaluate other LVLMs. 

%\vspace{-1mm}
\subsection{Ablation Studies}
%\vspace{-1mm}
\paragraph{(i) Effect of Reference Type:}
\label{ablation_reference}
In this section, we compare the performance variation of different LVLMs in reference-based and reference-free scenarios (see Table \ref{tab:judgement_accuracy_reference_comparison}). LVLMs that achieve more than 20\% pairwise judgment accuracy in OpenCQA are selected for the analysis. While we find that different LVLMs have a slight change in performance with the presence and absence of references, the performance difference between them based on a paired t-test is not statistically significant ($ p > 0.05 $). This demonstrates that the open-source LVLMs are robust in both reference-based and reference-free evaluation.  %find that the performance difference between them is not statistically significant ($p \leq 0.05$). This demonstrates that the open-source LVLMs are robust in both reference-based and reference-free evaluation. 

\paragraph{(ii) Effect of Evaluation Criteria:} In Table \ref{tab:error_distance_ablation_comparison}, we analyze the performance differences among various LVLMs with an error distance below $2.5$ in VisText (L1) across multiple evaluation metrics: (i) informativeness, (ii) relevance, and (iii) factual correctness. While we observe slight performance variations based on the evaluation criteria, the paired t-test demonstrates that these differences are not statistically significant ($p > 0.05$), indicating robust performance across various evaluation measures.

%\end{itemize}

\begin{table}[t]
\scriptsize
\setlength{\tabcolsep}{1.5pt} 
    \centering
    \begin{tabular}{l|c|c}
        \toprule
        \textbf{Model} & \textbf{With Reference} & \textbf{Without Reference} \\
        \midrule
                \rowcolor{open_models_below_4B!50} Qwen2-VL-2B-Instruct & 47.4 & 55.7 \\
                  %\rowcolor{open_models_below_4B!50} PaliGemma-3B-pt-224  & 40.22 & 45.50 \\
                   %\rowcolor{open_models_below_4B!50} ChartGemma-3B  & 11.96 & 8.66 \\
     \rowcolor{open_models_below_4B!50}Phi-3.5-Vision-3.8B-Instruct & 51.6 & 47.3 \\
  \rowcolor{open_models_below_4B!50} XGen-MM-Phi3-3.8B-Instruct & 66.8 & 68.4 \\
            \midrule
            \rowcolor{open_models_7B_12B!50} Janus-Pro-7B & 45.9 & 47.3 \\
            \rowcolor{open_models_7B_12B!50} Qwen2-VL-7B-Instruct & 66.7 & 67.8 \\
      \rowcolor{open_models_7B_12B!50} InternLM-Xcomposer2d5-7B  & 62.1 & 67.5 \\
              \rowcolor{open_models_7B_12B!50} LLaVA-Next-v1.6-Mistral-7B  & 71.0 & 73.0 \\
            \rowcolor{open_models_7B_12B!50} LLaVA-Critic-7B & \textbf{74.9} & \textbf{75.3} \\
       \rowcolor{open_models_7B_12B!50} mPLUG-Owl3-7B  & 63.5 & 58.2 \\
        \rowcolor{open_models_7B_12B!50} MiniCPM-V-2.6 & 63.2 & 65.4 \\
        \rowcolor{open_models_7B_12B!50} Idefics-9B-Instruct & 16.6 & 24.2 \\
        %\rowcolor{open_models_7B_12B!50} Llama-3.2-11B-Vision-Instruct & 5.37 & 7.42 \\
        \bottomrule
    \end{tabular}
    %\vspace{-1mm}
    \caption{Judgment Accuracy in comparison to GPT-4o in OpenCQA based on Reference-based (with reference) and Reference-free (without reference) evaluation.}
    \label{tab:judgement_accuracy_reference_comparison}
\end{table}

\begin{table}[t]
\tiny
\setlength{\tabcolsep}{1pt} 
    \centering
    \begin{tabular}{l|c|c|c}
        \toprule
        \textbf{Model} & \textbf{Factual Correctness} & \textbf{Informativeness} & \textbf{Relevancy} \\
        \midrule
                \rowcolor{open_models_below_4B!50} Qwen2-VL-2B-Instruct & 2.6 & 1.6 & 2.0 \\

     \rowcolor{open_models_below_4B!50} Phi-3.5-Vision-3.8B-Instruct  & 1.6 & 1.3 & 1.4 \\
  \rowcolor{open_models_below_4B!50} XGen-MM-Phi3-3.8B-instruct-r-v1 & 1.7 & 1.4 & 1.6 \\
            \midrule
            \rowcolor{open_models_7B_12B!50} Janus-Pro-7B & 1.4 & 1.0 & 1.3 \\
            \rowcolor{open_models_7B_12B!50} Qwen2-VL-7B-Instruct & {0.7} & {0.4} & {0.5} \\
      \rowcolor{open_models_7B_12B!50} InternLM-Xcomposer2d5-7B  & 1.0 & 0.8 & 0.9 \\
              \rowcolor{open_models_7B_12B!50} LLaVA-Next-v1.6-Mistral-7B   & 1.7 & 1.4 & 1.5 \\
            \rowcolor{open_models_7B_12B!50} LLaVA-Critic-7B & \textbf{0.6} & \textbf{0.3} & \textbf{0.4} \\
       \rowcolor{open_models_7B_12B!50} mPLUG-Owl3-7B  & 1.1 & 0.9 & 1.1 \\
        \rowcolor{open_models_7B_12B!50} MiniCPM-V-2.6 & 1.7 & 1.3 & 0.9 \\
        %\rowcolor{open_models_7B_12B!50} Idefics-9B-Instruct & 5.0 & 4.99 & 4.31 \\
        %\rowcolor{open_models_7B_12B!50} Llama-3.2-11B-Vision-Instruct & 3.99 & 3.63 & 4.27 \\
        \bottomrule
    \end{tabular}
    %\vspace{-1mm}
    \caption{Average Error Distance (compared with LLaVA-70B-Critic) in VisText (L1) for different LVLMs based on various Evaluation Types. Here, lower values indicate better performance.}
    \label{tab:error_distance_ablation_comparison}
\end{table}

%\vspace{-1mm}
\section{Conclusion and Future Work}
%\vspace{-1mm}
In this paper, we conducted a comprehensive evaluation of open-source LVLMs as automatic judges for chart comprehension and reasoning tasks. Our analyses revealed that while some open-source LVLMs (e.g., 7B models like LLaVA-Critic, Qwen2-VL, InternLM, and LLaVA-Next) can achieve judgment accuracy (with lower error rates) that is comparable to state-of-the-art closed-source models like GPT-4 or larger open-source models like LLaVA-Critic-70B; other models, such as ChartGemma and PaliGemma, struggle significantly, highlighting variability in their reliability. % Thus, the findings from our extensive experiments offer insights into building better-automated evaluators and chart-specific models in the future. 
Despite the promising results of various models, issues like bias and lack of instruction following capability still persist.  Therefore, future work should focus on mitigating biases, improving instruction following evaluation capability, alongside ensuring consistency across diverse evaluation criteria by developing a multimodal LLM judge using more recent models \cite{bai2025qwen2.5vl}. %
%tailored 
%for chart model evaluation. %Our code, judgment data, %(100k samples), 
%and 
%the proposed Chart-Instruct-Eval benchmark will be made publicly available. % upon acceptance.

%, providing a unified framework for seamless evaluation

\section*{Acknowledgments}
\vspace{-1mm}
This research is supported by the Natural Sciences and Engineering Research Council (NSERC) of Canada, the York Research Chairs (YRC) program, and Compute Canada. This paper is accepted at the \textbf{ACL 2025 Industry Track} and we thank all the anonymous reviewers for their excellent review comments. We would also like to thank Mehrad Shahmohammadi for his help in this research. 

\section*{Ethical Considerations}
\vspace{-1mm}
The models used for experiments are only used as the judge to evaluate other LVLM-generated responses. Therefore, 
the LVLM responses do not pose any ethical concerns. 
Additional compensation for human evaluation is not needed since it was conducted by two authors of this paper. 

% Since the focus of this work is real-world utilization, LVLMs that are available for both research and commercial purposes are only selected.

%\section*{Acknowledgments}
\vspace{-2mm}

\bibliography{custom}

\appendix
%\clearpage
\section{Appendix}
\label{sec:appendix}

\subsection{Regarding Model and Dataset Selection}
\label{appendix:data_model_selection_criteria}

%With the rapid development of numerous LVLMs recently, our evaluation may not cover many LVLMs. Nonetheless, we tried to use as many state-of-the-art LVLMs as possible (we mainly focused
We selected popular LVLMs that were released by early 2025, with sizes less than 10B parameters. %and conducted extensive experiments to investigate their effectiveness and limitations. 
%While some LVLMs that are proposed at the time of writing this paper may yield different results \cite{bai2025qwen2.5vl}, the findings from our study will still offer substantial insights into building better-automated evaluators and chart-specific LLM judges in the future. 
Although there are many other chart benchmarks currently available  \cite{huang2024pixels}, %(e.g., ChartQA \cite{masry-etal-2022-chartqa}, fact-checking over chart images \cite{akhtar-etal-2023-reading, akhtar-etal-2024-chartcheck}) require qualitative evaluation. Therefore, 
we selected OpenCQA and VisText since qualitative evaluation is often required in these datasets \cite{islam2024large}. %Moreover, we prioritize VisText over other chart captioning datasets since it contains both L1 and L2/L3 type captions. 

%\subsection{Performance based on Reference Type}
%\label{app:ablation_reference}

%In Table \ref{tab:judgement_accuracy_reference_comparison}, we compare the performance variation of different LVLMs that achieve more than 50\% pairwise judgment accuracy in OpenCQA in reference-based and reference-free scenarios. While we find that different LVLMs have a slight change in performance with the presence and absence of references, the performance difference between them is not statistically significant ($p \leq 0.05$) based on a paired t-test. This demonstrates that the open-source LVLMs are robust in both reference-based and reference-free evaluation. 

%\subsection{Performance based on Evaluation Type}
%\label{evaluation_reference}

%In Table \ref{tab:error_distance_ablation_comparison}, we compare the performance variation of different LVLMs that have error distance below 2.0 in VisText (L1) based on various evaluation criteria: (i) informativeness, (ii) relevance, (ii) factual correctness. We find that different LVLMs have a slight change in performance depending on the type of evaluation. Nonetheless, the performance difference between them based on a paired t-test is not statistically significant ($p \leq 0.05$). This demonstrates the robustness of open-source LVLMs.

\subsection{Prompts for the LVLM Judge}
\label{appendix_prompt}

%In this section, we demonstrate some sample prompts that we used for the LVLM judges.  

\definecolor{attachedColor}{HTML}{e0efff}
\definecolor{attachedColor2}{HTML}{f3f3f3}

\begin{tcolorbox}[
boxrule=0.25pt,   
  colback=attachedColor2,    % Main box background
  colframe=black,           % Black border
  colbacktitle=attachedColor, 
  coltitle=black,           % Black title text
  title={{OpenCQA Pointwise (With Reference)}},
  fonttitle=\bfseries,      % Bold font for the title
  fontupper=\small          % Smaller font for the box content
]

Suppose, you are a human evaluator. 
  You are required to rate the \{Evaluation Criteria\} of the answer generated by a model in comparison to the gold reference answer for a given question in the open-ended chart question answering task. \\

        Please provide your response in JSON format with the following keys: (i) Score, (ii) Explanation. \\

        Here, the 'Score' should be between 1 to 5 (inclusive), with the higher score indicating better. Moreover, the value for the "Explanation" key will contain the reason behind your scoring.  \\

        You should only provide the response in the required JSON format without any additional text. \\ % such that I can correctly parse the result from your JSON formatted response. \\

        In the following, you are first given the question, followed by the gold reference answer. 
        Afterward, you are given the model-generated answer. You are also provided with the chart image as the context for the chart question-answering task. \\

        [Question] \\

        [Gold Reference Answer] \\

        [Model Generated Answer] \\

        [\texttt{Chart Image}]

\end{tcolorbox}

\begin{tcolorbox}[
boxrule=0.25pt,   
  colback=attachedColor2,    % Main box background
  colframe=black,           % Black border
  colbacktitle=attachedColor, 
  coltitle=black,           % Black title text
  title={{OpenCQA Pairwise (Without Reference)}},
  fonttitle=\bfseries,      % Bold font for the title
  fontupper=\small          % Smaller font for the box content
]

Suppose, you are a human evaluator. 
         You are given the answers generated by two different models for a given question in the open-ended chart question answering task. Now, your task is to determine which model is better in terms of \{Evaluation Criteria\}. \\

        Please provide your response in JSON format with the following keys: (i) Model, (ii) Explanation, 
\\

        Here, the output value for the 'Model' key is the respective model that is better, could be either 'Model A' or 'Model B', or 'Tie' if both models are equally good. Moreover, the value for the "Explanation" key will contain the reason behind your preference.  
\\

        You should only provide the response in the required JSON format without any additional text. % such that I can correctly parse the result from your JSON formatted response.
\\

        In the following, you are first given the question. 
        Afterward, you are given the model-generated answers. You are also provided with the chart image as the context for the chart question-answering task. \\

        [Question] \\

        [Model 1 Generated Answer] \\

        [Model 2 Generated Answer] \\

        [\texttt{Chart Image}]

\end{tcolorbox}

\begin{tcolorbox}[
boxrule=0.25pt,   
  colback=attachedColor2,    % Main box background
  colframe=black,           % Black border
  colbacktitle=attachedColor, 
  coltitle=black,           % Black title text
  title={{VisText L1 Pointwise (With Reference)}},
  fonttitle=\bfseries,      % Bold font for the title
  fontupper=\small          % Smaller font for the box content
]

Suppose, you are an human evaluator.
You are required to rate the \{Evaluation Criteria\} of the L1 caption describing the aspects of the chart’s construction (e.g., chart type and axis labels) generated by a model in the chart captioning task. \\

Please provide your response in JSON format with the following keys: (i) Score, (ii) Explanation. \\

Here, the 'Score' should be between 1 to 5 (inclusive), with the higher score indicating better. Moreover, the value for the "Explanation" key will contain the reason behind your scoring. \\

You should only provide the response in the required JSON format without any additional text such that I can correctly parse the result from your JSON formatted response. \\

In the following, you are first provided with the gold reference caption. Afterward, you are given the model generated caption. You are also provided with the chart image which was used as the context for the chart captioning task. \\

        [Gold Reference Caption] \\

        [Model Generated Caption] \\

        [\texttt{Chart Image}]

\end{tcolorbox}

\begin{tcolorbox}[
boxrule=0.25pt,   
  colback=attachedColor2,    % Main box background
  colframe=black,           % Black border
  colbacktitle=attachedColor, 
  coltitle=black,           % Black title text
  title={{VisText L2/L3 Pairwise (No Reference)}},
  fonttitle=\bfseries,      % Bold font for the title
  fontupper=\small          % Smaller font for the box content
]

Suppose, you are a human evaluator.
You are given the captions generated by two different models in the chart captioning task. Now, your task is to determine which model is better based on \{Evaluation Criteria\}.  \\

Please provide your response in JSON format with the following keys: (i) Model, (ii) Explanation. \\

Here, the output value for the 'Model' key is the respective model that is better, could be either 'Model A' or 'Model B', or 'Tie' if both models are equally good. Moreover, the value for the "Explanation" key will contain the reason behind your preference. \\

You should only provide the response in the required JSON format without any additional text such that I can correctly parse the result from your JSON formatted response. \\

In the following, you are provided with the model generated captions. You are also provided with the chart image which was used as the context for the chart captioning task. \\

        [Model 1 Generated Caption] \\

        [Model 2 Generated Caption] \\

        [\texttt{Chart Image}]

\end{tcolorbox}

 % are also robust in terms of variations in evaluation criteria. 

\subsection{Error Analysis}
\label{error_analysis}

Some example error cases are demonstrated below.

% \begin{comment}
\begin{figure*}[t!]
    \centering
    \includegraphics[width=\textwidth]{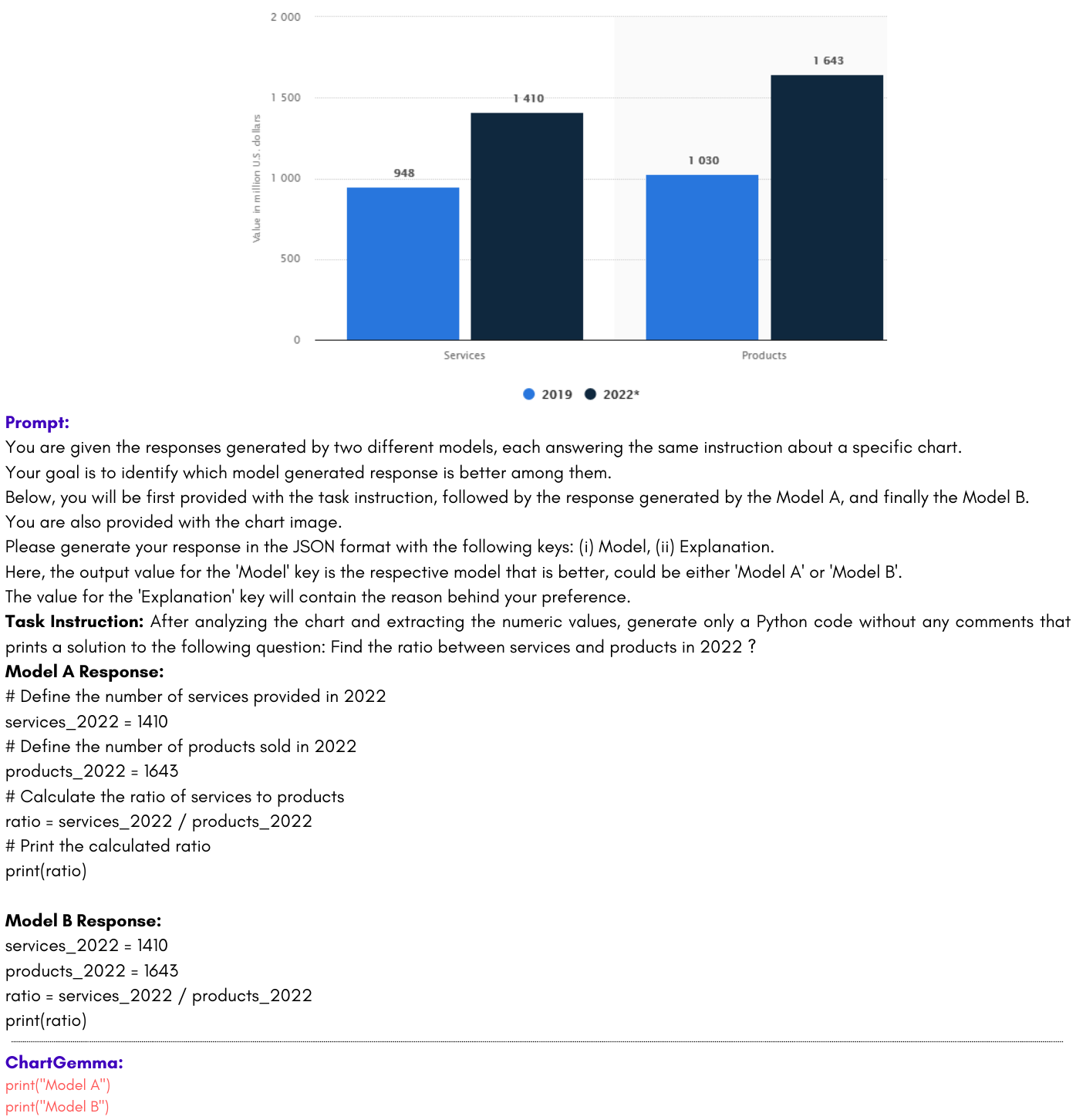}
    \caption{An example of an error case involves the ChartGemma model being tasked with evaluating the Python code responses generated by two different models and providing a verdict on which one is better. However, instead of following the instructions, the model failed to complete the task correctly and simply returned two print statements as its output (highlighted in \textcolor{red}{red text}).}
    %\vspace{-1mm}
    \label{fig:inst_follow_error}
\end{figure*}
% \end{comment}

\begin{figure*}[t!]
    \centering
    \includegraphics[width=\textwidth]{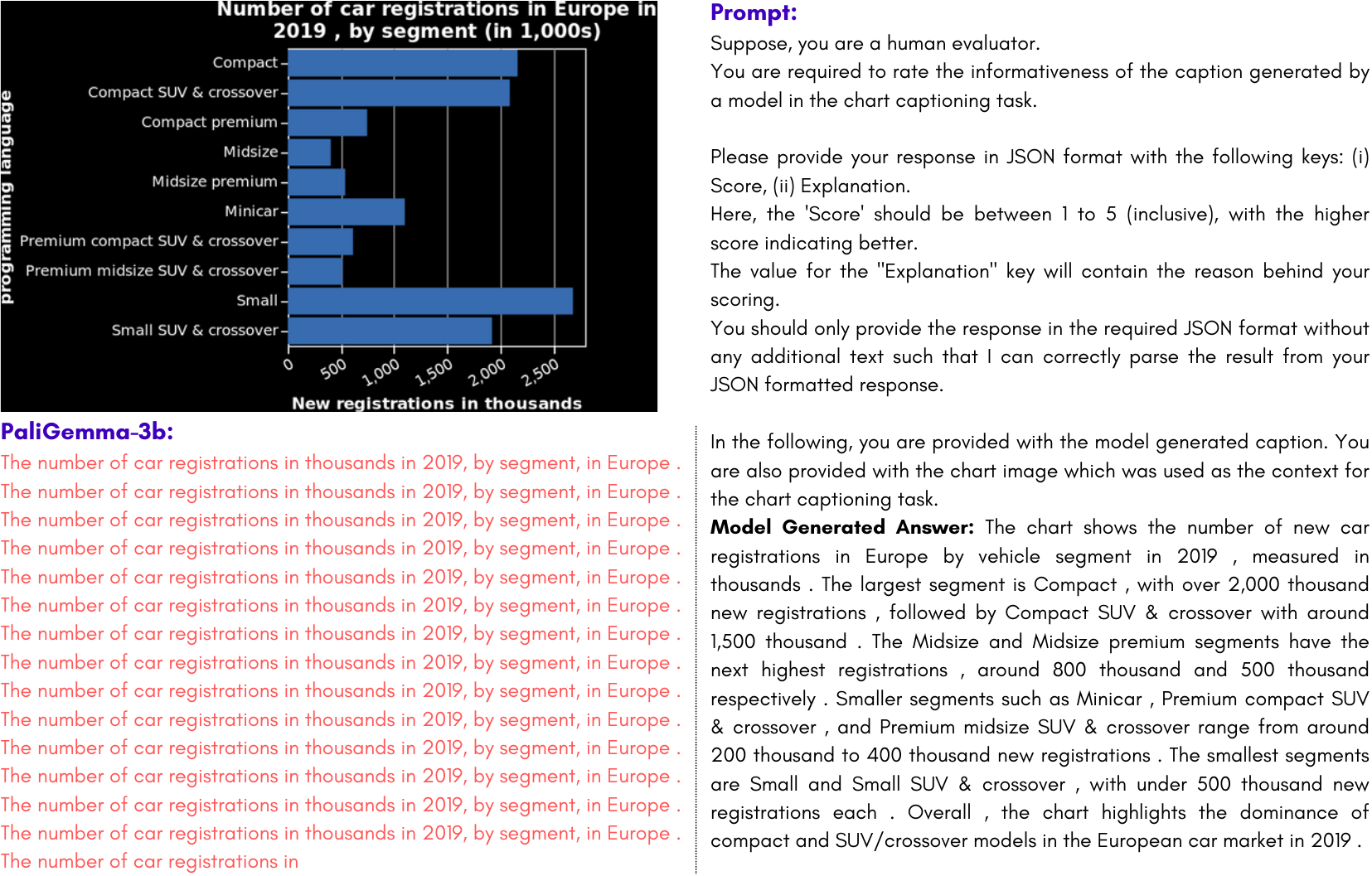}
    \caption{An example of an error case involves the PaliGemma-3b model being tasked with evaluating a chart caption generated by another model. Specifically, it was asked to rate the caption on a scale of 1 to 5 based on the `Informativeness' criterion and to provide an explanation for the rating. However, instead of performing the evaluation correctly, the model hallucinated and repeatedly generated the same line without adhering to the required JSON format. (highlighted in \textcolor{red}{red text}).}
    %\vspace{-1mm}
    \label{fig:form_follow_error}
\end{figure*}

\begin{figure*}[t!]
    \centering
    \includegraphics[width=\textwidth]{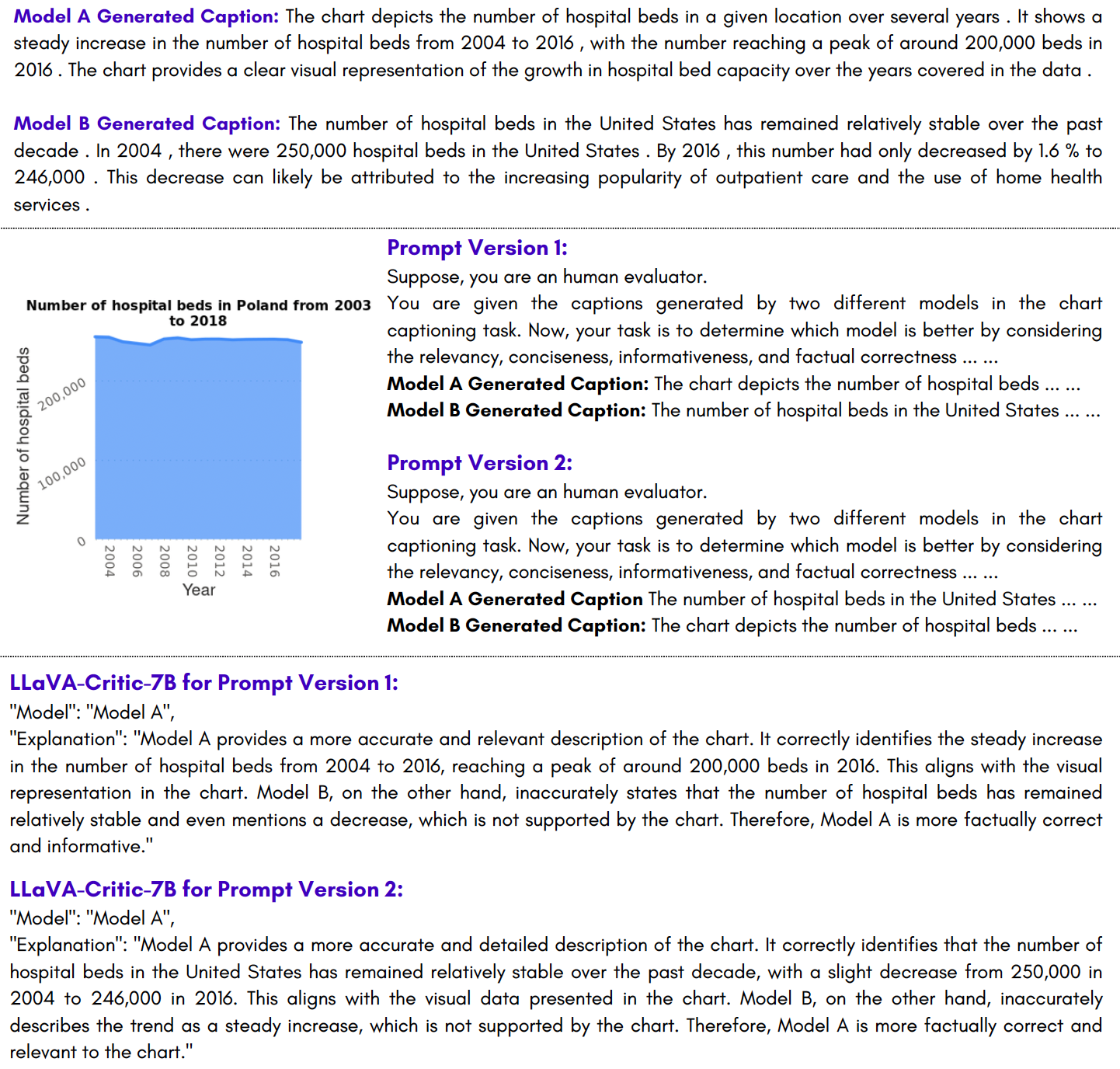}
    \caption{An example of an error case for the LLaVA-Critic-7B model which demonstrates position bias by changing its selection of the better caption based on the change in the order of the model generated captions.}
    %\vspace{-1mm}
    \label{fig:position_bias}
\end{figure*}

\begin{figure*}[t!]
    \centering
    \includegraphics[width=\textwidth]{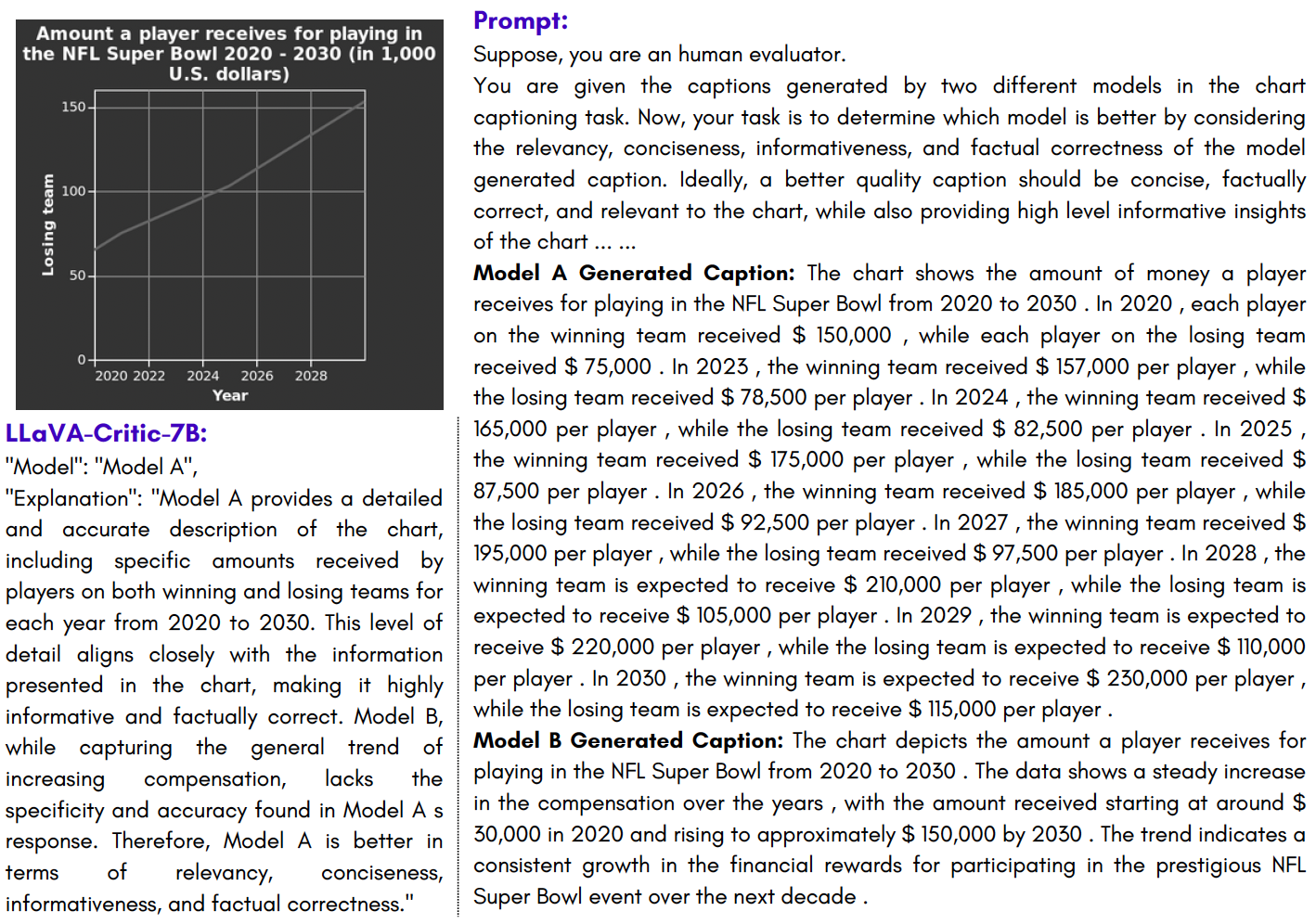}
    \caption{An example of an error case for the LLaVA-Critic-7B model which demonstrates length bias by selecting the lengthy caption even though in the given multidimensional prompt, conciseness was one of the criteria for a better caption.}
    %\vspace{-1mm}
    \label{fig:length_bias}
\end{figure*}

\end{document}